%% file: root.tex
\documentclass[letterpaper, 10 pt, conference]{ieeeconf}  

\IEEEoverridecommandlockouts                              

\usepackage{graphicx}
\usepackage{mathtools}
\usepackage{multirow}
\usepackage{amsmath}

\usepackage[backref]{hyperref} 
            
\usepackage{color}
\input{_color_note.tex}

\usepackage{cite}
\usepackage{amsmath,amssymb,amsfonts}
\usepackage{algorithmic}
\usepackage{graphicx}
\usepackage{textcomp}
\usepackage{xcolor}
\def\BibTeX{{\rm B\kern-.05em{\sc i\kern-.025em b}\kern-.08em
    T\kern-.1667em\lower.7ex\hbox{E}\kern-.125emX}}

\usepackage{diagbox}
\usepackage{algorithm}
\usepackage{algorithmic}

\usepackage{multirow}
\usepackage{array}

\usepackage{times}
\usepackage{epsfig}
\usepackage{graphicx}

\usepackage[switch]{lineno}
\usepackage{comment}
\usepackage{dsfont}
\usepackage{colortbl}
\usepackage[normalem]{ulem}

\newcommand{\PreserveBackslash}[1]{\let\temp=\\#1\let\\=\temp}
\newcolumntype{C}[1]{>{\PreserveBackslash\centering}p{#1}}
\newcolumntype{R}[1]{>{\PreserveBackslash\raggedleft}p{#1}}
\newcolumntype{L}[1]{>{\PreserveBackslash\raggedright}p{#1}}

\title{\LARGE \bf
GRM: Gradient Rectification Module for Visual Place Retrieval
}

\author{Boshu Lei$^{1}$ Wenjie Ding$^{2}$ Limeng Qiao$^{2}$ Xi Qiu$^{2}$ 
\thanks{*This work was done in MEGVII Inc. }
\thanks{$^{1}$Boshu Lei is with Faculty of Automation, Xi'an Jiaotong Univerisity,
        Xi' an, China}%
\thanks{$^{2}$ Wenjie Ding, Limeng Qiao and Xi Qiu are with MEGVII Inc.}%
}

\begin{document}

\maketitle
\thispagestyle{empty}
\pagestyle{empty}

\begin{abstract}

Visual place retrieval aims to search images in the database that depict similar places as the query image. However, global descriptors encoded by the network usually fall into a low dimensional principal space, which is harmful to the retrieval performance. We first analyze the cause of this phenomenon, pointing out that it is due to degraded distribution of the gradients of descriptors. Then, we propose Gradient Rectification Module~(GRM) to alleviate this issue. GRM is appended after the final pooling layer and can rectify gradients to the complementary space of the principal space. With GRM, the network is encouraged to generate descriptors more uniformly in the whole space. At last, we conduct experiments on multiple datasets and generalize our method to classification task under prototype learning framework.

\end{abstract}

\section{INTRODUCTION}

Visual place retrieval~(VPR) has gained much popularity in recent years. Given a query image, VPR aims to find images in the database that depict similar places. The typical pipeline of this task is to perform rough matching using global features and refine similarity using local features \cite{PatchNetVLAD,DELG,DELF}.  Most of the methods aim to obtain distinctive global and local features by improving the optimization function~\cite{GCL}, aggregation methods~\cite{NetVLAD,NetFV} and local branch structures~\cite{PatchNetVLAD,DELG}. However, few methods consider the feature distribution, which is very important for VPR task because it directly affects the retrieval results.


%


\RV{Early work to address the significance of feature distribution for retrieval is \cite{NegativeEvidence}. The author claims that whitening can balance out the co-occurrence of features, which is beneficial to retrieval. Later, many works try to eliminate the unbalanced distribution of descriptors, like loss regularizor \cite{Spread,UnderstandingContrastive} and normalization technique\cite{BatchNorm,ZCA,DeepRetrieval}. However, no one goes deeper into the question, what makes the model generate such unbalanced features. After thorough analysis, we attribute it to the gradients of descriptors in training process. During training, we find that both the descriptors and the corresponding gradients are trapped in the same low dimensional principal space. As updates to the descriptors, gradients trapped in principal space won’t push descriptors to other dimensions, resulting in the network generating degraded features and thus harmful to the retrieval performance. We term the span of the eigenvectors with large eigenvalues from feature covariance matrix as principal space and the span of remaining vectors as complementary space.}


    \begin{figure}[t]
		\vspace{6pt}
		\centering
		\includegraphics[width=8.7cm]{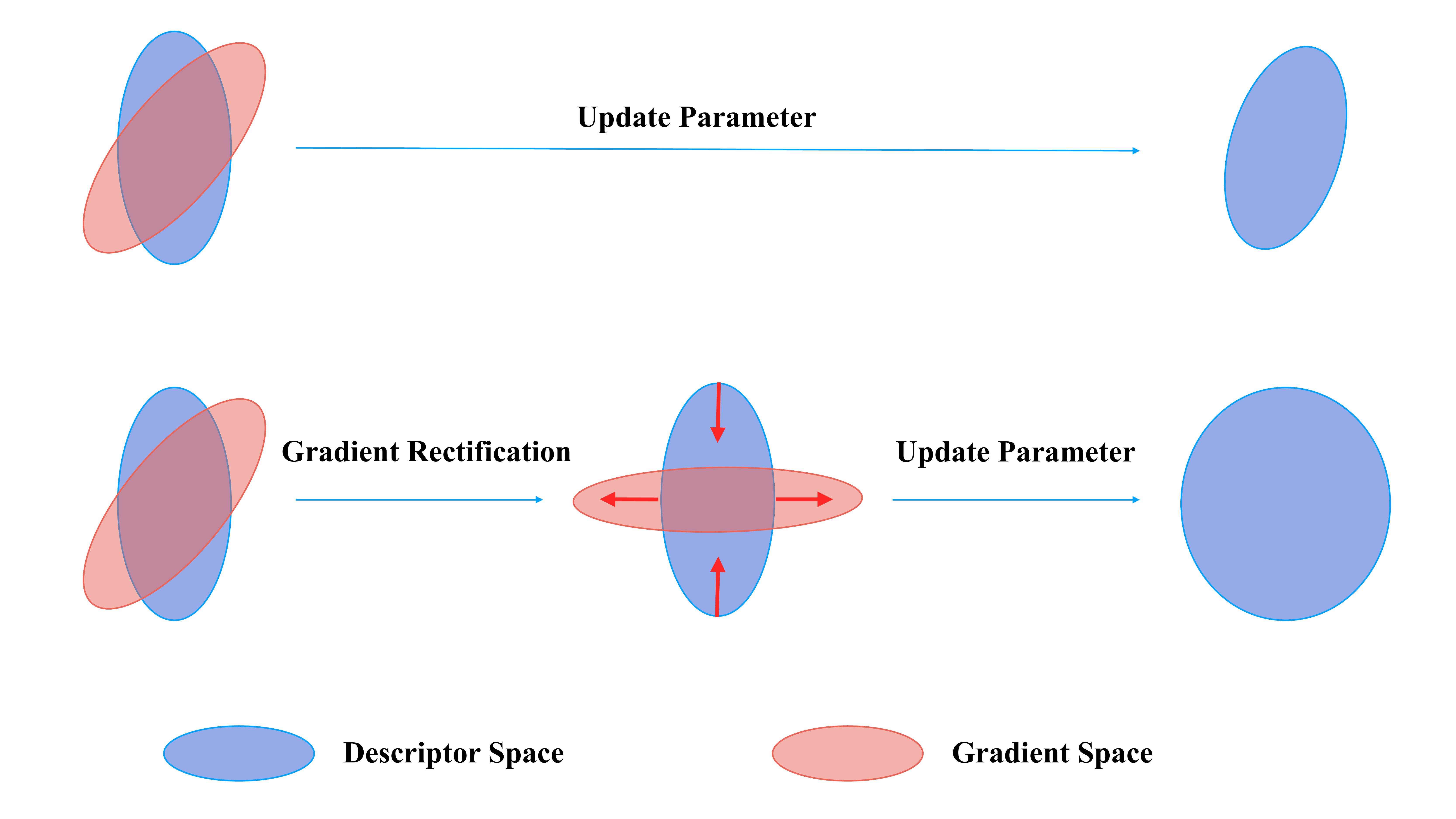} \\
		\caption{\textbf{Gradient Rectification.} \RV{The circles in the figure represent the distribution of descriptors and gradients respectively.} \WJ{ Without GRM}, network generates degraded features when gradients fall into the same principal space of descriptors~(Up). When GRM is applied, gradients are pulled out of this space (in orthogonal direction) and the network will generate descriptors more uniformly (Down). }
	\label{fig:motivation}
	\end{figure}


\RV{We now propose to modify the distribution of descriptors through a \textbf{G}radient \textbf{R}ectification \textbf{M}odule (GRM) as illustrated in Fig.~\ref{fig:networks}.} First, we propose a new memory queue method to estimate the co-variance matrix of descriptors. This covariance matrix is used to generate a projection matrix to rectify gradients. Second, we propose our rectification criterion. We inhibit the component of gradients along the principal space and stimulate the component along directions orthogonal to the principal space. After doing so, gradients will be deviated from the principal space and descriptors are pulled to the complementary space of the principal space. In the forward pass, GRM stores descriptors in the memory queue for covariance estimation. In the backward pass,  GRM modifies gradients based on the covariance matrix estimated from queue. It's worth noting that GRM is plug-and-play and parallel to those methods that aim to improve the loss function or local branches.

    \begin{figure*}[th]
		\flushleft
		\small
		\centering
		\includegraphics[width=16cm]{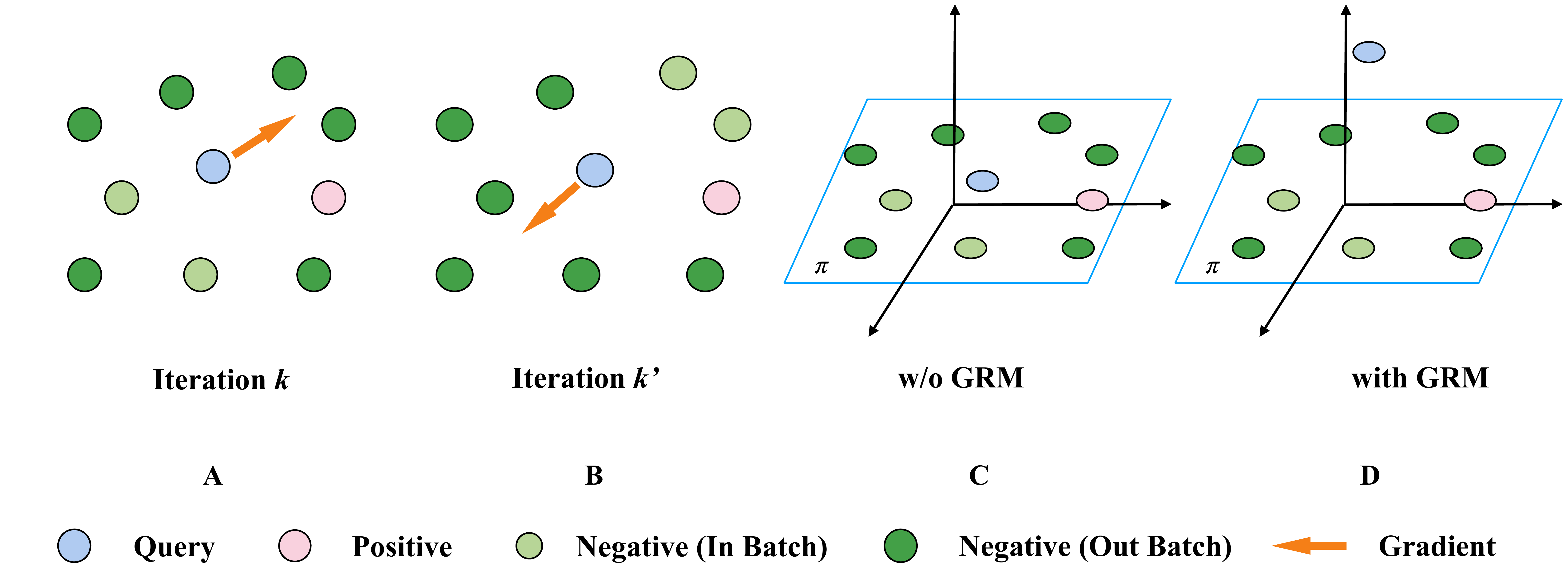} \\
		\caption{\textbf{Optimization dilemma in low dimensional space.} In current step (A), the influence from the negative and positive instances will push the query instance to the direction along the arrow. In later steps (B), when negative instances in the opposite direction are sampled, they will push gradients back. We propose to pull the query out of the principal space ((C) to (D)). In that case, the query will be further away from all negative samples. }
	\label{fig:zigzag}
	\end{figure*}
	

In conclusion, our contributions are threefold:

1)~We point out that gradients fall into the same space of descriptors \WJ{in VPR task}, hindering the network from exploring other dimensions of the feature space.

2)~We propose a simple yet effective Gradient Rectification Module~(GRM). By modifying the gradients of descriptors, GRM pulls descriptors to the complement space of the principal space and alleviates the problem above.

3)~We conduct extensive experiments in various datasets in VPR and classification tasks. Experimental results show that the proposed method significantly outperforms state of the arts. 

\section{Related Works}

\subsection{Global Feature Descriptor}

Early approaches to extract global feature descriptor are based on aggregation of hand crafted local features. Methods are like Bag of Words (BoW)~\cite{BoW} and Vector of Locally Aggregated Descriptors (VLAD)~\cite{VLAD}. Sparse keypoint location or dense sampling on image are applied to aggregate local features. When deep learning technology is introduced into these frameworks, new methods like NetVLAD~\cite{NetVLAD}, NetBoW~\cite{NetBoW}, NetFV~\cite{NetFV} emerge. Other methods apply simple pooling layer, eg. max pooling or GeM pooling~\cite{GeM}, on top of the CNN architecture~\cite{DeepRetrieval,GCL}. These methods are trained with Triplet Loss~\cite{NetVLAD}, or ranking-based loss~\cite{APLoss,Rank}. Most recently,~\cite{GCL} proposes to use a continuous measure to calculate the similarity between images. Instead of depending on the binary relation, they use the calculated similarity score to train the neural network and achieve huge success in MSLS challenge~\cite{MSLS}.

    \begin{figure*}[th]
		\centering
		\includegraphics[width=18.0cm]{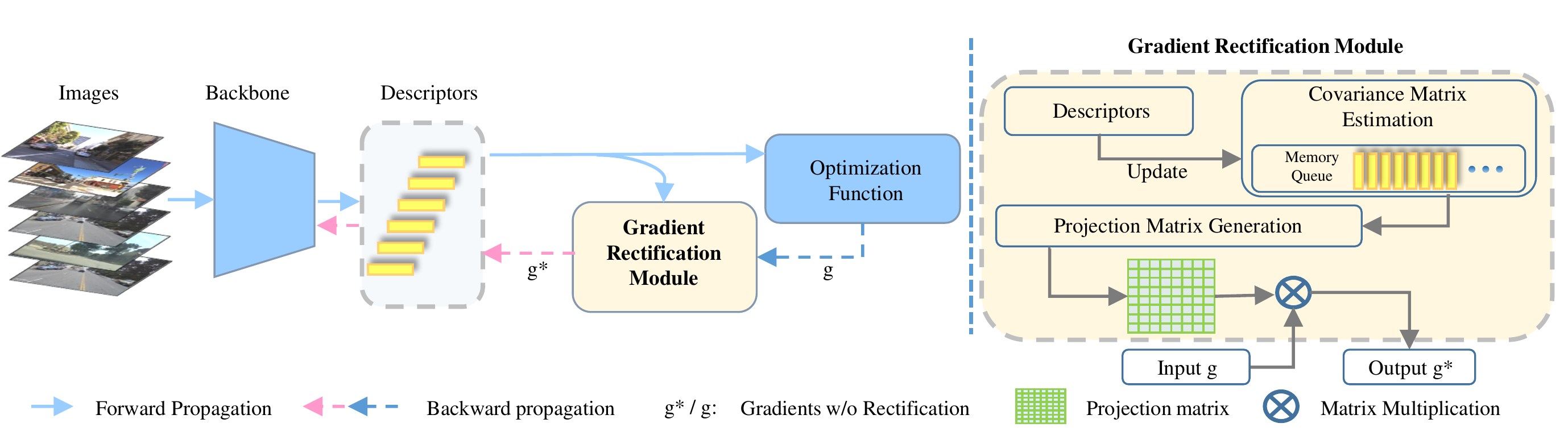} \\
		\caption{\textbf{The learning paradigm of the proposed method. } Forward descriptors are accumulated in the memory queue and later used to generate the projection matrix. Backward gradients are multiplied by this matrix when flowing through the GRM. } 
	\label{fig:networks}
	\end{figure*}

\subsection{Local Feature Refinement}

Early methods for refinement are separated from global feature matching scheme. Local features like SIFT\cite{SIFT}, SURF\cite{SURF} are extracted and matched based on nearest neighbor criterion. Then model-based methods, such as RANSAC\cite{RANSAC} and PROSAC\cite{PROSAC} are used to generate transformation hypotheses based on correspondences. The final retrieval result is re-ranked by the number of inliers under the hypothesis. DELF\cite{DELF} proposes to use the feature map before pooling for later refinement process. PatchNetVLAD\cite{PatchNetVLAD} proposes to extract NetVLAD\cite{NetVLAD} descriptors on image patches for later refinement procedure.\cite{DELG} is the first to jointly learn local and global features for retrieval task.

\subsection{Whitening and Spreading Methods}

\LBS{According to~\cite{NaturalNet}, when the features become white, the Fisher information matrix reduces to an identity matrix. \LBS{In that case,} the simple stochastic gradient descent optimization coincides with the natural gradients descent optimization which is more suitable for difficult optimzation tasks. Whitening methods can be classified into two categories. The first category is to add extra penalty terms in loss function. Zhang et al.~\cite{Spread} proposed to add a regularizor to spread vectors uniformly in the feature space. In~\cite{UnderstandingContrastive},  the author provides a thorough analysis of the uniformity and alignment metric of the feature distribution space. Additional loss penalties are added to adjust the uniformity and alignment of feature space. Another method is to process the features directly, either in the forward pass or the backward pass. Batch Normalization (BN)~\cite{BatchNorm} is the most widely used method in the forward pass of deep learning model. Later,~\cite{ZCA} constructs the whitening matrix based on zero-phase component analysis to mitigate the stability issue in BN and PCA~\cite{PCAIssue}. \cite{DeepRetrieval} appends a PCA projection layer behind the network to spread descriptors. ~\cite{GradTransform} finds that the whitening matrix exerted on features can be instead applied on model weights. They prove that their methods are mathematically equivalent to ZCA~\cite{ZCA} but the computation complexity is largely reduced. WADAM~\cite{WADAM} extends the idea in \cite{GradTransform} with momentum, adaptive dampening and proposes new optimizers for network training. } 

In VPR, \cite{NegativeEvidence} points out that whitening can balance out the co-occurrence of features, which is generally beneficial to retrieval tasks. Whitened features can boost the performance of the network on the current datasets and improve the generalization ability of the network across datasets~\cite{GCL}. However, Whitening adds a strong prior on the distribution of the descriptors and is not involved in the training process. 
 

\LBS{
Our work focuses on the global descriptor extraction procedure, since the quality of the global descriptor is the bottleneck of later local descriptor matching process. Unlike previous works in VPR ~\cite{GCL,APLoss,TransVPR,NetVLAD}, we do not propose new aggregation methods or new loss functions, but instead focus on the feature distribution that lacks attention. We propose a new method to optimize the feature distribution, which is parallel to \WJ{most} VPR methods before-head. 
}

\section{Method}

\subsection{Gradient Distribution Problem}

We claim that pair-wise loss function in VPR is one of the reasons for the degraded distribution of gradients. A general pair-wise loss function can be written as

\begin{equation}
L = \sum_{i} f(s_{i, 1}, ..., s_{i, N})
\end{equation}

where $s_{i, j}$ is the similarity between query $i$ and sample $j$. It is usually defined as cosine similarity or L2 distance. These two are equivalent when the descriptors are normalized. For simplicity, we will use the L2 distance.

\begin{equation}
s_{i,j} = ||\vec{p}_i - \vec{p}_j||^2 , \quad \vec{p}_i \in R^C
\end{equation}

$\vec{p}_i$ and $\vec{p}_j$ are the descriptors encoded by the network. The gradient of the descriptor is 

\begin{equation}
\frac{\partial L}{\partial \vec{p}_i} = \sum_{j=1}^{N} \frac{\partial f}{\partial s_{i, j}} 2 (\vec{p}_i - \vec{p}_j)
\label{eq:partial}
\end{equation}

Since $2\frac{\partial f}{\partial s_{i, j}}$ is a scalar, Eq. \ref{eq:partial} can be written as 

\begin{equation}
\frac{\partial L}{\partial \vec{p}_i} = \sum_{j=1}^{N} \alpha_{j} (\vec{p}_i - \vec{p}_j)
\end{equation}

\RV{Here, we use contrastive loss as an example. The contrastive loss (CL) is shown in Eq. \ref{eq:CL}}.

\begin{equation}
    L_{CL}(\vec{p_i}) = \sum_{j=1}^{N}\phi_{i,j} s_{i, j} + (1-\phi_{i, j}) max(\tau - s_{i,j}, 0)
    \label{eq:CL}
\end{equation}

$\phi_{i,j} = 1$ if sample i and j are positive pair and 0 otherwise. $\tau$ is the threshold. The derivative is 

\begin{equation}
    \frac{\partial L}{\partial \vec{p}_i} = \sum_{j=1}^{N} [\phi_{i,j} - (1 - \phi_{i,j}) sign(\tau - s_{i,j})] (\vec{p}_i - \vec{p}_j)
    \label{eq:example}
\end{equation}

\RV{In Eq. \ref{eq:example}, since $\phi_{i,j} - (1 - \phi_{i,j}) sign(\tau - s_{i,j})$ is a scalar, the gradient is a linear combination of the query and database descriptors.} Since the projection of descriptors in the principal space is much larger than the projection in the complement space, a linear combination of these principal components is generally larger than other components. Therefore, gradients will also have large components in the principal space. 


\subsection{Gradient Rectification Module}
The whole process can be summarized in Fig.~\ref{fig:networks}. During training, we feed images into the network to get descriptors. These descriptors are used to update the estimated covariance matrix $P$. Loss is computed according to the learning task, for instance, GCL~\cite{GCL} or Triplet Loss. Then we backward the loss and capture gradients at the descriptor level. The projection matrix $P^*$ is applied here and the modified gradients flow into the network.

\subsubsection{Covariance Matrix Estimation}
A brutal way to estimate the covariance matrix is to feed all training examples into the network and calculate the covariance matrix with them after each epoch. This can be applied when the size of the dataset is relatively small. However, when the size of the dataset is large, this operation is time consuming. What is worse, since the descriptors are usually designed to be high dimensional, the number of examples inside one batch is inadequate to provide an accurate estimation of the covariance matrix, if using the regular formula below:

\begin{equation}
P = \frac{1}{B-1} \sum_{i=1}^{B}(\boldsymbol{x} - \bar{\boldsymbol{x}})(\boldsymbol{x} - \bar{\boldsymbol{x}})^T.
\end{equation}

Here, $P$ is the estimated covariance matrix. $B$ is the batch size. $\boldsymbol{x}$ is the descriptor and $\bar{\boldsymbol{x}}$ is the mean descriptor within the current batch.

Therefore, non-singularity and positive-definite property of the covariance matrix may not be preserved. To this end, we borrow the idea, memory queue, from unsupervised learning domain~\cite{Moco}. We manage a queue of size K. The current batch is enqueued to the memory queue and the oldest batch in the queue is removed. The queue represents a sampled subset of all the instances. Covariance matrix are estimated using the standard form from instances in memory queue. \LBS{
    In order to prevent numerical instability, we also add a small number ($10^{-3}$) to each of  the diagonal elements of P.
}
\subsubsection{Projection Matrix Generation}

We apply eigenvalue decomposition on $P$: 

\begin{equation}
P = U diag(\lambda_1, ..., \lambda_n) U^T
\end{equation}

$\lambda_1, ..., \lambda_n$ are eigenvalues in descending order. Then we form the projection matrix $P^*$ to modify gradients.

\begin{equation}
P^* = U \mathop{diag}(\frac{\bar{\lambda}}{\lambda_1}, ..., \frac{\bar{\lambda}}{\lambda_n})^{s} U^T
\end{equation}

$\bar{\lambda}$ is the mean of the eigenvalues. To eliminate the scale problem, we put $\bar{\lambda}$ instead of a constant on the numerator. Since at the initial stage of the training process, the elements in the descriptors are all near zero, causing all the eigenvalues to be very small. A constant in the numerator may cause a large scaling value along this dimension, leading to unstable training at the beginning. \WJ{$s$ is a hyper-parameter controlling the rectification rate}. \LBS{
Large $s$($>1$) causes big rectifying value and leads to unstable training while small $s$($<0.5$) is unable to push the network to generate descriptors uniformly. We manually set this parameter to 1.
}

\subsubsection{Gradient Rectification} We rectify gradients using projection matrix $P^*$ as follows

\begin{equation}
g^* = P^* g
\end{equation}

Modified gradients $g^*$ flows into the backbone network. Next, optimizers like SGD or Adam\cite{Adam} are adopted to update network's parameters. In the ideal case, when the covariance matrix of  gradients is only different from the covariance matrix of the descriptors by a constant multiplier, it is trivial to show that the variance of gradients is identical along each dimension.

\RV{The space complexity of GRM is $(K+C)C$, where $C$ is the dimension of descriptors. $KC$ is the size of the memory queue and $C^2$ is the size of the covariance matrix. The time complexity for covariance matrix estimation is $O(KC^2)$ and for eigenvalue decomposition is $O(C^3)$. GRM is not involved in the inference stage, it doesn't impact inference speed of the model.}


\subsection{Datasets}

\subsubsection{MSLS} MSLS is a large scale place recognition dataset that contains images taken in 30 different cities~\cite{MSLS}. The training set contains over 500k query images and 900k map images. The validation set consists of 19k map images and 11k query images and the test set has 39k map images and 27k query images. Two images are considered as similar if they are taken by cameras located within 25m of distance and with less than 40$^{\circ}$ of viewpoint variation.

\subsubsection{Pittsburg} This dataset contains images of urban environments gathered from google street view in the city of Pittsburgh, Pennsylvania, USA~\cite{NetVLAD}. We use the test set of Pitts30K to test our model. The test set of Pitts30K contains 6k query images and 10k map images. 

\subsubsection{Tokyo 24/7} It consists of images taken in Tokyo, Japan, with large variations of illumination, as they are taken during day and night~\cite{Tokyo}. The query set consists of 315 images and the map contains 76k photos.

\subsubsection{Nordland} This dataset is recorded on a train at four different seasons~\cite{Nordland}. Each image is taken at the same place as corresponding ones in other seasons. We use the test subset, which consists of 3.5k images in each season, to test our model. The tolerance for positive match is set to 10 frames and we use images in summer as queries and images in winter as databases, as PatchNetVLAD\cite{PatchNetVLAD}.

\subsection{Implementation Details}

We use GCL Loss~\cite{GCL} to train our model. We select ResNet50~\cite{ResNet} and ResNext101~\cite{ResNext} as backbone to extract features. GeM pooling is applied to get the global descriptor. The memory queue size (K) is 10,240. We use Adam optimizer\cite{Adam} and set the learning rate to 1e-4. We train our model for 200 epochs~(10k pairs in MSLS per epoch). Following GCL~\cite{GCL}, we use the weights pretrained on ImageNet and  only train the last two blocks of the backbone. For GRM, the rectification rate is set to 1. We train our network on MSLS training set and test them on other datasets. Top-N recall~(R@N) is adopted as evaluation metric.


\begin{table*}[ht]
		\centering\scriptsize
		\caption{\textbf{Comparison with State-of-the-art Methods.}}
		\resizebox{1.0\textwidth}{!}{
			\renewcommand{\arraystretch}{1.2}
			\begin{tabular}[c]{ C{3.0cm}  C{0.45cm} C{0.45cm}  C{0.45cm}  C{0.45cm} C{0.45cm} C{0.45cm}  C{0.45cm}  C{0.45cm}  C{0.45cm} C{0.45cm} C{0.45cm} C{0.45cm} C{0.45cm}  C{0.45cm}  C{0.45cm}}
			\hline
			\hline
				\multicolumn{1}{c}{\multirow{2}{*}{Recall~(\%)}} & 
				\multicolumn{3}{c}{\textbf{MSLS (Test. set)}} &
				\multicolumn{3}{c}{\textbf{MSLS (Val. set)}} & 
				\multicolumn{3}{c}{\textbf{Pittsburgh 30k}} & 
				\multicolumn{3}{c}{\textbf{Nordland}} &
				\multicolumn{3}{c}{\textbf{Tokyo247}}\\
				\cline{2-16}
				& R@1 & R@5 & R@10 & R@1 & R@5 &  R@10 & R@1 & R@5 & R@10 & R@1 & R@5 & R@10 & R@1 & R@5 & R@10 \\
				\hline
				\hline
			NetVLAD 64~\cite{NetVLAD} & 45.1 & 58.8 & 63.7 & 70.1 & 80.8 & 84.9 & 68.6 & 84.7 & 88.9 & - & - & - & 34.0 & 47.6 & 57.1 \\
			\hline
			NetVLAD 16~\cite{NetVLAD} & 39.4 & 53.0 & 57.5 & 70.5 & 81.1 & 84.3 & 70.3 & 84.1 & 89.1 & - & - & - & 37.8 & 53.3 & 61.0 \\
			\hline
			Patch-NetVLAD~\cite{PatchNetVLAD} & 
			48.1 & 57.6 & 60.5 & 79.5 & 86.2 & 87.7 & \textbf{87.7} & \textbf{94.5} & \textbf{95.9} & 71.3 & 80.9 & 82.2 & \textbf{86.0} & \textbf{88.6} & \textbf{90.5} \\
			\hline
			GCL~(ResNet50)~\cite{GCL} & 52.9 & 65.7 & 71.9 & 74.6 & 84.7 & 88.1 & 79.9 & 90.0 & 92.8 &  44.7 & 58.8 & 65.3 & 58.7 & 71.7 & 76.8 \\
			\hline
			GCL~(ResNext101)~\cite{GCL} & 62.3 & 76.2 & 81.1 & 80.9	& 90.7	& 92.6 & 79.2 & 90.4 & 93.2 & 69.9 & 85.6 & 90.1 & 58.1 & 74.3 & 78.1 \\
			\hline
			TransVPR w/o. rerank~\cite{TransVPR} & 48.0 & 67.1 & 73.6 & 70.8 & 85.1	& 89.6 & 73.8 & 88.1 & 91.9 & - & - & - & - & - & - \\
			\hline
			\hline
			\textbf{Ours}~(ResNet50) & 62.3 & 75.5 & 78.7 & 82.4 & 90.2 & 91.7 & {82.8} & {91.9} & 93.9 & 66.2 & 89.9 & 93.0 & 55.9 & 70.8 & 76.2 \\
			\hline
			\textbf{Ours}~(ResNext101)& \textbf{64.9} & \textbf{77.2}	& \textbf{81.1} & \textbf{82.6} & \textbf{90.7} & \textbf{92.6}  & 81.3 & 91.4 & 93.5 & \textbf{82.3} & \textbf{92.4} & \textbf{95.2} & 62.5 & 75.2 & 82.5 \\
			
			\hline
			\hline
			\end{tabular}
		}
		\label{table:SOTA}
	\end{table*}

\subsection{Comparison with State-of-the-art Methods.}
	
For a fair comparison, we only compare the first stage of all methods without the re-ranking process. We compare the performance of GRM with state of the arts in Table~\ref{table:SOTA}. On MSLS dataset, even though the baseline methods, GCL-ResNet50 and GCL-ResNext101, are superior than all the other methods, our methods can still boost the performance. ResNext101 trained by our method can achieve the best R@5 in the official MSLS challenge, at 77.2\%. For ResNet50, the R@5 increases nearly 10\%, and is comparable to the larger ResNext101 model. Our methods can also bring huge improvement on Nordland dataset. This is due to the fact that season variation consists a large portion of the MSLS dataset. Therefore, our methods trained on MSLS generalize well on Nordland dataset, increasing over 5\% on ResNext101 and over 30\% on ResNet50. In urban environment, Pittsburg 30K and Tokyo24/7, our method can bring moderate improvement on the baseline method, between 1\textasciitilde2\%, compared to the baseline method. Our best R@5 on Pittsburg30K (91.9\%) and Tokyo24/7 (75.2\%) is lower than PatchNetVLAD (94.5\% on Pittsburg30K and 88.6\% on Tokyo24/7). This is because Patch-NetVLAD is fine-tuned on Pittsburg30K training dataset, with images including drastic variations in illumination. However, in order to keep aligned with other methods, we only train our method on MSLS dataset. 
	
\subsection{Comparison with Other Whitening Methods}

\LBS{
    We compare the performance of GRM and other feature whitening methods. We select GCL~\cite{GCL} as our baseline and use multiple strategies to expand the feature space. For Spread Loss~\cite{Spread}. We set the weight for the regularization term to 1. For AP Loss~\cite{APLoss}. We use the GCL Label in~\cite{GCL} to sample 2 positive images and 7 negative images. Negative images are half hard negative (similarity  score greater than 0 but less than 0.5) and half negative (similarity score equals 0). We compare our method to~\cite{WADAM}. Standard Adam~\cite{Adam} optimizer is replaced by WADAM ~\cite{WADAM} in the GCL~\cite{GCL} framework.}
    
    
    The training procedure follows the previous section and we show the R@N index on MSLS val and test sets in Table \ref{table:whiten}. GRM exceeds all methods in Table \ref{table:whiten}. The performance drops when adding the extra regularization in~\cite{Spread}. We conjecture that the inclusion of regularization might break the original distribution and leads to a sub-optimal distribution. As we can see, AP Loss~\cite{APLoss}, as the representation of multi-pair form, increase R@5 by 4\% on MSLS test set but is still lower than our method. Meanwhile, AP Loss suffers from large memory consumption and long training time. As for the WADAM~\cite{WADAM}, it can bring slight improvement on MSLS val., but when evaluated on test set, it under-performs the baseline method.

\LBS{
    The result corroborates that it is neither the weak repulsive force nor the limited batch size that causes the unbalanced distribution of descriptors but the phenomenon that gradients and descriptors fall into the same low dimensional principal space. Since gradients won't push the descriptors out of this principal space, descriptors will dwell in this principal space and never be uniformly distributed.
}


    \begin{table}[ht]
		\centering\scriptsize
		\caption{\textbf{Comparison with Whitening Method.}}
		\resizebox{0.48\textwidth}{!}{
 			\renewcommand{\arraystretch}{1.2}
			\begin{tabular}[c]{ C{2.3cm} C{0.4 cm} C{0.4 cm} C{0.4 cm} C{0.4 cm} C{0.4 cm} C{0.4 cm}}
			\hline
			\hline
			
				\multicolumn{1}{c}{\multirow{2}{*}{Recall~(\%)}} & 
				\multicolumn{3}{c}{\textbf{MSLS (Test. set)}} &
				\multicolumn{3}{c}{\textbf{MSLS (Val. set)}} \\
				\cline{2-7}
				& R@1 & R@5 & R@10 & R@1 & R@5 &  R@10 \\
				\hline
				\hline
			ResNet50~\cite{GCL} & 52.9 & 65.7 & 71.9 & 74.6 & 84.7 & 88.1 \\
			\hline
			Spread Loss~\cite{Spread} & 30.8 & 47.2 & 53.6 & 49.1 & 63.4 & 68.6 \\
			\hline
			AP Loss~\cite{APLoss} & 53.4 & 69.7 & 75.0 & 72.3 & 82.4 & 84.9  \\
			\hline
			WADAM~\cite{WADAM} & 48.4 & 62.6 & 68.7 & 72.9 & 85.1 & 87.9  \\
			\hline
			\textbf{ResNet50 + GRM} & \textbf{62.3} & \textbf{75.5} & \textbf{78.7} & \textbf{82.4} & \textbf{90.2} & \textbf{91.7} \\
			\hline
  			\hline
			\end{tabular}
		}
		\label{table:whiten}
	\end{table}

\vspace{.5cm}

    \begin{figure}[ht]
		\small
		\centering
		\includegraphics[width=8.6cm]{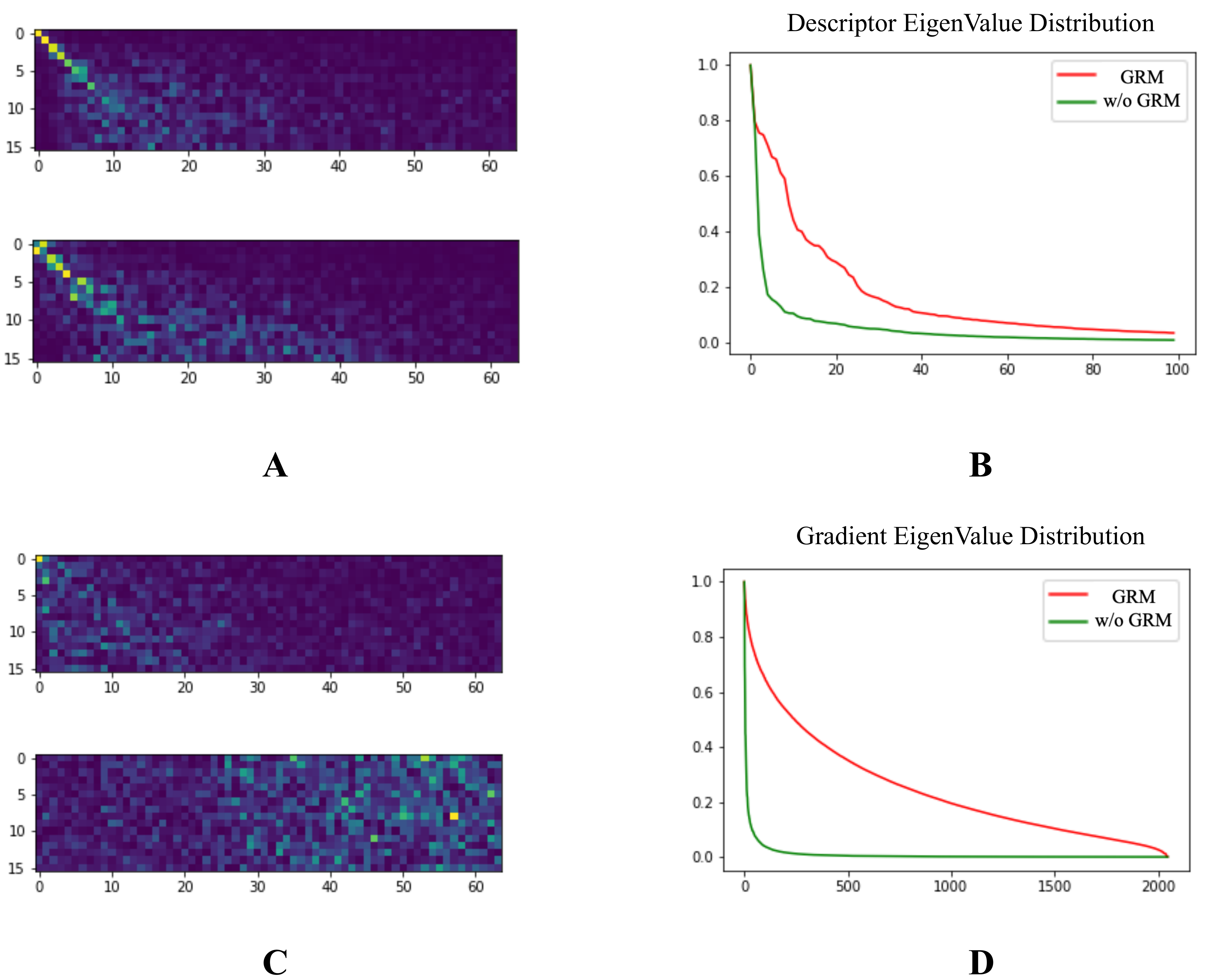} \\
		\caption{\textbf{Gradient and descriptor distribution.} (A) and (C) are alignment of eigenvectors between different training stage w/o. GRM and with GRM. (B) and (D) are the eigenvalue distribution of descriptors and gradients with and w/o. GRM.}
	\label{fig:gradient}
	\end{figure}

\subsection{Ablation Study}

\subsubsection{Gradient distributions analysis}
\label{sec:distributionvis}


\LBS{
In order to illustrate the problem that the descriptors encoded by the network fall into a low dimensional space and do not change during the training process, we plot the correlation between the principal space of descriptors in different stages during training process in Fig.~\ref{fig:gradient}. In the upper figure of (A), pixel value at (i, j) is the dot product between the i-th eigenvetor of the feature space encoded by model after training 30 epochs and the j-th eigenvector of the feature space encoded by model after training 90 epochs, \textbf{without GRM}. Here, all the eigenvectors are sorted by their corresponding eigenvalue. The i-th eigenvector is the eigenvector corresponds to the i-th largest eigenvalue. Therefore, the brighter the pixel is, the more aligned two eigenvectors are. In the lower figure of (A), pixel value at (i, j) is the dot product between the i-th eigenvetor of the feature space and the j-th eigenvector of the gradient space. Fig.~\ref{fig:gradient} (C) is the same as (A) but the model is trained \textbf{with GRM}. 
}

\LBS{
In Fig.~\ref{fig:gradient} (A), We notice that brightest pixels are on the diagonal line of the image, indicating that eigenvectors are aligned respectively during training process and the principle space does not change. The same situation remains between the principle space of descriptors and gradients. Given that descriptors and gradients all concentrate in a low principal space (the green curve in Fig.~\ref{fig:gradient} B and D), we claim that descriptors trapped in a low dimensional space is the result of the gradients which fall into the same dimensional space. 
}

When GRM is applied, bright pixels disperse in Fig.~\ref{fig:gradient} (C). Hence the principal space between descriptors and gradients are not aligned. Gradients are pulled out of the principal space of descriptors and influence the distribution of descriptors in later steps. We also plot the eigenvalue of the descriptor space and the gradient space in Fig.~\ref{fig:gradient} (B) and (D). We can find that the descriptors and gradients distribute more uniformly when GRM (red curve) is applied.

\subsubsection{Experiments in Classification Tasks.} To evaluate the performance of the proposed method on classification tasks, we test our method on CIFAR-10, CIFAR-100, Caltech 101 and Caltech 256 datasets~\cite{Caltech}. We adopt the prototype learning method in~\cite{PrototypeLearning}. We use ResNet18 as backbone to extract features and use average pooling to get the descriptor. The feature dimension is set to 512. We use the GCPL loss in~\cite{PrototypeLearning} to train our network. The gradient rectification module is applied to descriptors and prototypes simultaneously. The model is optimized using SGD with momentum of 0.9. The initial learning rate is set to 0.05 and is multiplied by 0.7 every 20 epochs. We initialize all prototypes as zero~\cite{PrototypeLearning} to avoid noise. The whole training process lasts for 200 epochs. We use the top-1 accuracy to evaluate all the models. Results are summarized in Table \ref{table:Classification}.	
	
GRM can bring improvement in prototype learning framework. It brings around 0.5\% improvement on CIFAR-10 and CIFAR-100 using the GCPL framework. On Caltech dataset, even though the GCPL framework~\cite{PrototypeLearning} underperforms the baseline method, our method can still achieve better results, about 1\textasciitilde2\% than the baseline method.

\begin{table}[ht]
		\centering\scriptsize
		\caption{\textbf{Performance on classification task.}}
		\resizebox{0.5\textwidth}{!}{
			\renewcommand{\arraystretch}{1.0}
			\begin{tabular}[c]{ C{2.8cm} C{1.1cm} C{1.3cm} C{1.1cm} C{1.1cm}}
			\hline
			\hline
			
			{\textbf{Accuracy~(\%)}} & 
			{\textbf{CIFAR-10}} &
			{\textbf{CIFAR-100}} & 
			{\textbf{Caltech101}} & 
			{\textbf{Caltech256}}\\

				\hline
				\hline
			ResNet18 (Baseline) & 93.2 & 73.3 & 74.3 & 73.5  \\
			\hline
			ResNet18 + GCPL & 94.4 & 75.5 & 71.5 & 71.4 \\
			\hline
			ResNet18 + GCPL + \textbf{GRM} & \textbf{94.9} & \textbf{75.8} & \textbf{75.2} & \textbf{75.3} \\
			\hline
			\hline
			\end{tabular}
		\label{table:Classification}
		}
	\end{table}

\subsubsection{Comparison with Other Estimation Method}

Another way to estimate the covariance matrix is running average. The procedure is given below:

\begin{equation}
\begin{aligned}
    N_{k+1} &= N_{k} + b \\ 
    \bar{x}_{k+1} &= \frac{N_{k+1}-b}{N_{k+1}} \bar{x}_{k} + \frac{b}{N_{k+1}} \sum_{i=1}^{b} x_i \\
    P_{k+1} &= \frac{N_{k+1}-b}{N_{k+1}} P_k + \frac{b}{N_{k+1}} \sum_{i}^{b} (x_i - \bar{x}_{k+1}) (x_i - \bar{x}_{k+1})^T 
\end{aligned}
\end{equation}

$b$ is the batch size. $k$ is the number of step. $x$ is the descriptor. $P$ is the co-variance matrix. $\bar{x}$ is the running mean. $P$ is initialized as identity and $\bar{x}$ is initialized as zero. In each step, the projection matrix is calculated from $P_{k}$ and multiplies the backward gradients. We use the same setting to test the running average method and compare it with baseline and memory queue method.

The running average method can still outperform the baseline method on MSLS. It gets a top-5 recall rate of 73.2\% on MSLS test set, higher than the baseline (65.5\%), which is still significant. On MSLS validation set, it gets top-5 recall at 88.4\%, 4\% higher than the baseline method. It is even slightly better on Pittsburg30K than the memory queue method. Moreover, running average method doesn't need to maintain a large memory queue to store descriptors, cutting off the memory consumption. 

\begin{table}[!th]
		\centering\scriptsize
		\caption{\textbf{Comparison with Running Average Method.}}
		\resizebox{0.5\textwidth}{!}{
			\renewcommand{\arraystretch}{1.2}
			\begin{tabular}[c]{ C{2.7cm}  C{0.6cm} C{0.6cm}  C{0.6cm}  C{0.6cm} C{0.6cm} C{0.6cm}}
			\hline
			\hline
				\multicolumn{1}{c}{\multirow{2}{*}{\small method}} & 
				\multicolumn{3}{c}{\textbf{MSLS (Test. set)}} &
				\multicolumn{3}{c}{\textbf{MSLS (Val. set)}} \\
				\cline{2-7}
				& R@1 & R@5 & R@10 & R@1 & R@5 &  R@10 \\
				\hline
				\hline
				
			GCL~(ResNet50)~\cite{GCL} & 52.9 & 65.7 & 71.9 & 74.6 & 84.7 & 88.1 \\
			\hline
			ResNet50~+~Average~(Sqrt)~ & 61.5 & 73.2 & 77.2 & \textbf{84.0}  & 88.7 & 90.7 \\
			\hline
			ResNet50~+~Bank~(Linear)~ & \textbf{62.3} & \textbf{75.5} & \textbf{78.7} & 82.4 & \textbf{90.2} & \textbf{91.7} \\
			
			\hline
			\hline
			\end{tabular}
		}
		\label{table:estimation}
	\end{table}

 \subsubsection{Memory Queue Size}

Here we conduct experiments to analyze the effect of memory queue size K. The result is summarized in Table \ref{table:queue}. We find that small queue will harm the performance of the model. This is because the estimated covariance matrix is not accurate from few samples. Queue size larger than 10,000 suffices an accurate estimation.  

 \begin{table}[!th]
		\centering\scriptsize
		\caption{\textbf{Memory Queue Size Comparison}}
		\resizebox{0.5\textwidth}{!}{
			\renewcommand{\arraystretch}{1.2}
\begin{tabular}[c]{ C{2.7cm}  C{0.6cm} C{0.6cm}  C{0.6cm}  C{0.6cm} C{0.6cm} C{0.6cm}}
			\hline
			\hline
				\multicolumn{1}{c}{\multirow{2}{*}{\small Memory Queue Size}} & 
				\multicolumn{3}{c}{\textbf{MSLS (Test. set)}} &
				\multicolumn{3}{c}{\textbf{MSLS (Val. set)}} \\
				\cline{2-7}
				& R@1 & R@5 & R@10 & R@1 & R@5 &  R@10 \\
				\hline
				\hline

                GCL~(ResNext101)~\cite{GCL} & 62.3 & 76.2 & 81.1 & 80.9	& 90.7	& 92.6 \\
                \hline
                K=5120 & 59.1 & 71.4 & 75.7 & 80.4 & 87.6 & 89.9 \\
                \hline 
			K=7680 & 62.4 & 73.8 & 76.8 & 81.9 & 89.1 & 90.6 \\
			\hline
			\textbf{K=10240} & \textbf{64.9} & \textbf{77.2}	& \textbf{81.1} & \textbf{82.6} & \textbf{90.7} & \textbf{92.6} \\
			\hline
                \hline
			\end{tabular}
		}
		\label{table:queue}
	\end{table}

\section{Conclusions}

In this article, we point out the gradient problem in training deep neural network for global feature extraction in VPR. The degraded distribution of gradients leads to the unbalanced distribution of descriptors. We propose our new GRM and memory queue method to rectify gradients. This method can improve the existing state of the art methods on global descriptor extraction. We also demonstrate that our method could be generalized to classification task. For future work, the GRM Module can be applied to other metric learning tasks like unsupervised learning or re-ID in tracking task. 

\bibliographystyle{IEEEtran}
\bibliography{IEEEexample}

\end{document}

%% file: _color_note.tex
\newcommand\LBS[1]{\textcolor{black}{#1}}
\newcommand\RV[1]{\textcolor{black}{#1}}

\newcommand\WJ[1]{\textcolor{black}{#1}}